\documentclass[10pt,twocolumn,letterpaper]{article}

\usepackage{cvpr}
\usepackage{times}
\usepackage{epsfig}
\usepackage{graphicx}
\usepackage{amsmath}
\usepackage{amssymb}
\usepackage{multirow}
\usepackage[ruled,vlined]{algorithm2e}
\usepackage{url}

\usepackage{authblk}
\usepackage{tabularx}
\usepackage{color}
\usepackage{siunitx}
\usepackage{wrapfig}
\usepackage{flexisym}

\usepackage[pagebackref=true,letterpaper=true,colorlinks,bookmarks=false,breaklinks]{hyperref}

\cvprfinalcopy 

\def\cvprPaperID{3} 

\ifcvprfinal\fi
\begin{document}

\title{All-In-One Underwater Image Enhancement using Domain-Adversarial Learning}

\author[1]{Pritish Uplavikar}
\author[1]{Zhenyu Wu}
\author[1]{Zhangyang Wang}
\affil[1]{Texas A\&M University, College Station, TX}
\affil[ ]{\textit {\{pritishuplavikar,\ wuzhenyu\_sjtu,\ atlaswang\}@tamu.edu}}

\maketitle

\begin{abstract}
   Raw underwater images are degraded due to wavelength dependent light attenuation and scattering, limiting their applicability in vision systems. Another factor that makes enhancing underwater images particularly challenging is the diversity of the water types in which they are captured. For example, images captured in deep oceanic waters have a different distribution from those captured in shallow coastal waters. Such diversity makes it hard to train a single model to enhance underwater images. In this work, we propose a novel model which nicely handles the diversity of water during the enhancement, by adversarially learning the content features of the images by disentangling the unwanted nuisances corresponding to water types (viewed as different domains). We use the learned domain agnostic features to generate enhanced underwater images. We train our model on a dataset consisting images of 10 Jerlov water types \cite{jerlov}. Experimental results show that the proposed model not only outperforms the previous methods in SSIM and PSNR scores for almost all Jerlov water types but also generalizes well on real-world datasets. The performance of a high-level vision task (object detection) also shows improvement using enhanced images with our model.
\end{abstract}

\section{Introduction}

Underwater images have an application in a variety of fields like marine research and underwater robotics. We need clear underwater imagery to study deteriorating coral reefs and other aquatic life. Underwater robotic systems also rely heavily on high quality images to fulfill their objectives. However, the quality of the images acquired for these applications is degraded due to various factors. One of the major factors for this degradation is wavelength dependent light attenuation over the depth of the object in the scene. For example, red light is absorbed in water at a higher rate than blue or green light. Hence, we see a blueish or a greenish tint in an underwater scene. Another factor diminishing underwater image quality is the light scattered due to the small particles present in water, which introduces a homogeneous background noise to the image.



\begin{figure}[t]
\centering
\includegraphics[width=.15\textwidth]{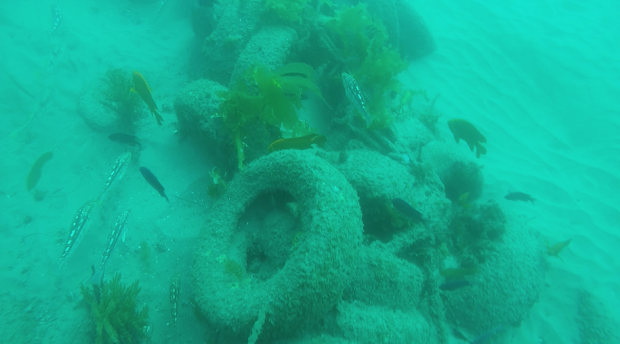}\hfill
\includegraphics[width=.148\textwidth]{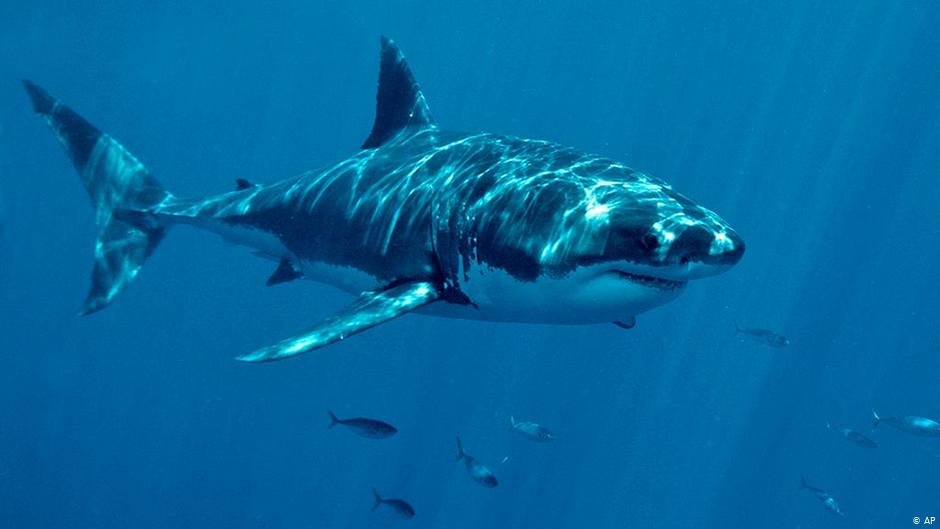}\hfill
\includegraphics[width=.152\textwidth]{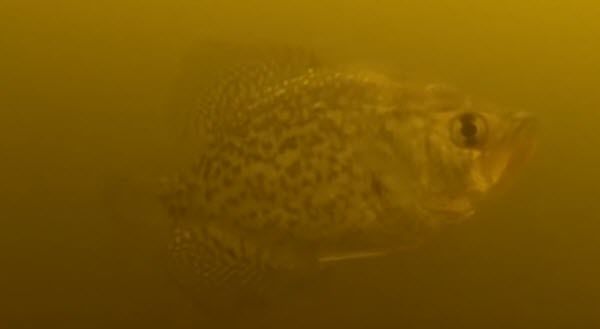}

\caption{Diversity of underwater scenes. Images are captured in (from left to right) coastal water, deep oceanic water and muddy water. Reprinted from \cite{img1}, \cite{img2} and \cite{img3} respectively.}\label{divcomp}
\end{figure}

 Apart from these factors, another challenge in underwater image enhancement is the diversity of underwater image distributions. We can see this diversity in figure \ref{divcomp}, which shows how underwater scenes captured in shallow coastal waters look different than those captured in deep oceanic waters or those captured in muddy waters. It is hard for a single model to enhance underwater images for such multiple image distributions and, therefore, providing a universal solution for underwater image enhancement is difficult. While previous work has addressed the challenges of light attenuation and scattering, not many have handled the challenge of image distribution diversity explicitly. \cite{uwhazelines} proposes color restoration of underwater images by performing color correction using attenuation coefficient ratios for all the Jerlov water types and then selecting the best result out of them. Whereas, \cite{anwar} proposes one solution by training multiple models, each for a different Jerlov water type. But these approaches seem inefficient and rely on the prior knowledge of the water type for the given image to perform color restoration.

One more challenge faced in underwater image enhancement is the lack of real-world datasets containing the ground truth clear images, as it is extremely difficult to find degraded and clear versions of the same real-world underwater scene, thus creating a bottleneck for data-driven methods. To address this challenge, synthetic underwater image datasets have been built in the past. One such example is the work done in \cite{anwar}, who synthesize an underwater image dataset consisting images of 10 Jerlov water types using the NYU Depth Dataset V2 \cite{nyudepth} which provides the ground truth clear images and the depth, both of which are required to synthesize degraded underwater images using the underwater image formation model \cite{imgformmodel}.

The task of enhancing underwater images is, therefore, difficult and has its own unique challenges. These images fail at many vision tasks like object detection, classification, segmentation which provokes a need to process them and enhance their quality. We propose a novel solution to this problem by addressing all the challenges mentioned above using a convolutional neural network \cite{cnn} based encoder-decoder to reconstruct clear underwater images and a convolutional neural network based classifier to classify the Jerlov water types, which acts as our \textit{nuisance} classifier. We synthesize a dataset of synthetic underwater images by following \cite{anwar} to train our model, thus taking into account the factors of wavelength dependent light attenuation and scattering of light due to small water particles while forming an underwater image and thereby addressing the respective challenges. To address the challenge of underwater image distribution diversity, we train our model to learn the domain agnostic features for a given degraded underwater image, where the domain is the Jerlov water type of the image. The objective of our encoder, apart from learning an encoding to reconstruct a clear underwater image, is to make the prediction of the nuisance classifier as uncertain as possible by discarding the features denoting the water type and preserves only the scene related features \cite{wu2018towards}. This is similar to a generator fooling a discriminator in a generative adversarial network \cite{gan}, the only difference being that in this case we want to make the classifier uncertain. We, therefore, introduce an adversarial loss on our encoder, computed at the end of the nuisance classifier, which is negative entropy. The encoder is also acted upon by a reconstruction loss computed at the end of our decoder. Thus, we propose a novel model which performs underwater color restoration for multiple types of underwater images.

Our approach has multiple highlights: i) Our proposed approach is able to learn water-type agnostic features. We adapt the adversarial training strategy proposed in \cite{wu2018towards} to our model; ii) Our proposed model outperforms the previous enhancement methods in SSIM and PSNR scores for almost all Jerlov water types; iii) Our model has good generalization ability on real-world datasets, as well as performs nicely on improving subsequent object detection on enhanced images.

\section{Related Work}

Many previous attempts to solve the underwater image enhancement problem have used physics-based methods. \cite{Jordt3d} tries to solve this problem by explicitly modeling the refraction in water, whereas, \cite{JaffeOUIS} incorporates the inherent properties of the underwater medium such as attenuation, scattering, and the volume scattering function in order to simulate image formation. \cite{imgformmodel} defines an underwater image formation model which is given as

\begin{equation} \label{imgform}
    U_{c}(x) = I_{c}(x)T_{c}(x) + B_{c}(1-T_{c}(x)), c \in \{r, g, b\}
\end{equation}

where $U_{c}(x)$ is a point $x$ in the underwater image, $I_{c}(x)$ is a point $x$ in the clear image, $T_{c}(x)$ is the fraction of the light reaching the camera after reflecting from point $x$ in the scene and $B_{c}$ is the homogeneous background light of the scene. $T_{c}(x)$ is further given as
\begin{equation} \label{transcoeff}
    T_{c}(x) = 10^{-\beta_{c}d(x)} = \frac{E_{c}(x,d(x))}{E_{c}(x,0)} = N_{c}d(x)
\end{equation}

where $\beta_{c}$ is the wavelength dependent medium attenuation coefficient, $E_{c}(x,d(x))$ is the energy of a light beam from point $x$ after it passes through a medium and $N_{c}(d(x))$ is the normalized residual energy ratio for every unit of depth covered.

The above physical model is similar to that of image dehazing, except that the medium attenuation coefficient is wavelength dependent, whereas in dehazing it does not depend on the light wavelength. This model has been used by many approaches to solve the underwater image enhancement problem. \cite{akkaynak} tries to improve on the above model by computing attenuation coefficients in the 3D RGB space, whereas \cite{anwar} uses the above model to generate a synthetic dataset of 10 Jerlov water types. We generate a similar dataset in our work, the details of which are given in section \ref{synds}.  

In recent years, deep learning \cite{dlbook} techniques like Convolutional Neural Networks (CNN) \cite{cnn} and Generative Adversarial Networks (GAN) \cite{gan} have been very effective at solving vision problems. Naturally, these techniques have then been used for underwater image enhancement. \cite{fabbrigan} trains a GAN to learn the mapping from underwater to clear images. \cite{anwar} train multiple CNN models, each for different water type in their dataset, to get enhanced images. However, these methods fail to provide a singular solution capable of handling the diversity of underwater images apart from generating their clear versions.

\section{Method}



Since one of our goals, apart from underwater image enhancement, is to train a single model which can do this task for multiple water types, we first try to learn a water type agnostic encoding for the given underwater image. That means, ideally, the latent vector $Z$ extracted from an encoder $E$ for the same underwater scene, should be the same for different water types. That way the decoder or the generator $G$ is able to reconstruct a clear image of the scene from only the scene specific features. Both $E$ and $G$ are neural networks in our model.

\begin{figure}[t]
\begin{center}
  \includegraphics[width=1\linewidth]{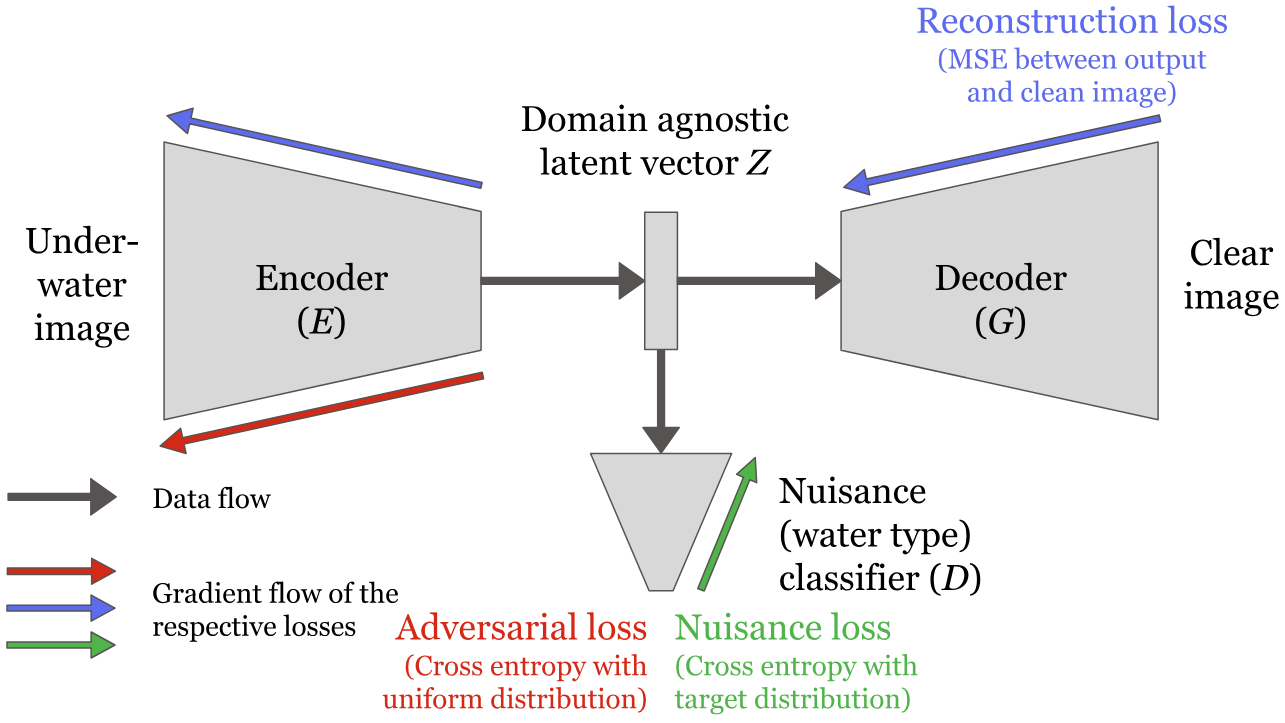}
\end{center}
  \caption{Our model architecture.}\label{arch}
\end{figure}

To do so, we introduce a novel application of a nuisance classifier $D$ along with $E$ and $G$. The nuisance classifier is a neural network which aims to classify the water type of the given input image from its latent vector $Z$ extracted from the encoder. However, we also introduce an adversarial loss \cite{gan} over the encoder using the nuisance classifier. Our formulation of the adversarial loss forces the encoder to generate $Z$ such that the nuisance classifier is unsure of the possible water types. Thus, the adversarial loss forces the encoding to be agnostic of the features denoting the water type. The full architecture can be seen in figure \ref{arch}.

\subsection{Losses}

Our model consists of three losses: the reconstruction loss $L_{R}$, the nuisance loss $L_{N}$ and the adversarial loss $L_{A}$. They force the model to generate a clear image while discarding the features denoting the water type. Detailed information about all the three losses can be found below.

\subsubsection{Reconstruction loss}

We compute a reconstruction loss $L_{R}$, which is the mean squared error between the image generated by $G$, from the latent vector $Z$, and the clear image ground truth $Y$ for the given input image $X$. The reconstruction loss is given as
\begin{equation}
    L_{R}(X, Y) = \frac{1}{N}.\sum_{i=1}^{N} \lvert G(Z)_{i} - Y_{i} \rvert ^2
\end{equation}
where $Z = E(X)$ and $N$ is the number of pixels.

\subsubsection{Nuisance loss}

We compute a nuisance loss $L_{N}$, which is the cross entropy with the target distribution of water types for the predicted distribution of water types from the nuisance classifier $D$, for the latent vector $Z$ of the input image $X$ of water type $C$. This nuisance loss is backpropagated to only update the nuisance classifier $D$. The nuisance loss is given as
\begin{equation}
    L_{N}(X, C) = -\sum_{c=1}^{M} y_{c} \log{D(Z)_{c}},
\end{equation}
where $y_{c} = 1$ if $c = C$ else $y_{c} = 0$, $Z = E(X)$ and $M$ is the number of classes.
\subsubsection{Adversarial loss}

As we want to increase the uncertainty or entropy of the nuisance classifier, we try to reduce the certainty or negative entropy of the classifier prediction. We, thus, compute an adversarial loss $L_{A}$, which is the negative entropy of the predicted distribution of water types from the nuisance classifier $D$, for the latent vector $Z$ of the input image $X$. This adversarial loss is backpropagated to only to update the encoder $E$. The adversarial loss is given as
\begin{equation}
    L_{A}(X) = \sum_{c=1}^{M} D(Z)_{c} \log{D(Z)_{c}}
\label{neeqn}
\end{equation}
where $Z = E(X)$ and $M$ is the number of classes.

\subsection{Training procedure}

 We first train only our encoder and decoder till a certain threshold, defined by the performance of the model on the validation set. We do this step to make sure that the encoder outputs an encoding $Z$ with meaningful features before we include the nuisance classifier in our model. We then train our model by following a procedure which prioritizes the adversarial training of the encoder, while also making sure that the nuisance classifier is strong enough. Keeping the nuisance classifier strong is critical for good adversarial training of the encoder. Algorithm \ref{tp} shows the training procedure we follow.
 
 \section{Experiments}

We train our model on the synthetic underwater image dataset described in detail in section \ref{synds}. The model is trained on a machine with the following configuration - Intel i7 6700 HQ processor, 8 GB RAM, NVIDIA GeForce GTX 960M 4GB graphics card.

\begin{algorithm}
\KwData{Encoder $E$, decoder $G$ and nuisance classifier $D$, $threshold_{G} \gets 0.9$, $threshold_{D} \gets 0.85$}
 Get $val_{G} \gets$ Cross validation SSIM score of $G$ \\
 \While{$val_{G} < threshold_{G}$}{
  Update $E$ and $G$ using $L_{R}$
 }
 \For{$n$ training epochs}{
 \uIf{$val_{G} < threshold_{G}$}{
   Update $E$ using $L_{R}$ and $L_{A}$, $G$ using $L_{R}$
   }\uElseIf{$val_{D} < threshold_{D}$}{
   Update $D$ using $L_{N}$
   }\Else{
   Update $E$ using $L_{R}$ and $L_{A}$, $G$ using $L_{R}$
  }
 Get $val_{G} \gets$ Cross validation SSIM score of $G$ \\
 Get $val_{D} \gets$ Cross validation accuracy of $D$
 }
 \caption{Training procedure of our model}
 \label{tp}
\end{algorithm}


\subsection{Model architectures}

\begin{figure}[h]
	\centering
	\includegraphics[width=1\linewidth]{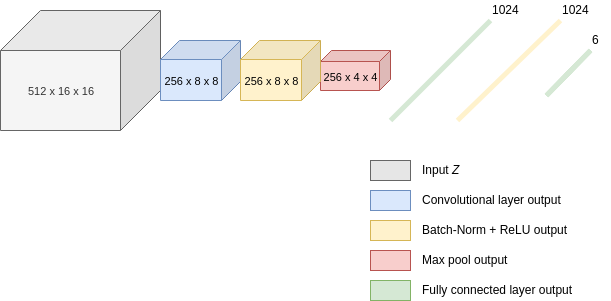}
	\caption{Our nuisance classifier architecture.}\label{nc}
\end{figure}

We use the architecture of U-Net \cite{unet} for our encoder-decoder. U-Net is useful as the skip connections between encoder and decoder provide local and global information for decoder to generate clear images from. Also, it is a fully convolutional neural network which means it can handle images of varying sizes. Our nuisance classifier is a convolutional neural network which predicts probability of 6 classes. Its architecture can be seen in figure \ref{nc}. 


\subsection{Datasets}

Previous methods have tried to synthesize degraded underwater images from their clear versions. We train our model on such a synthesized dataset built using the method described in \cite{anwar}. In order to see the usability of our model, we also test our model on a real-world dataset.

\subsubsection{Synthetic underwater image dataset\label{synds}}

\begin{figure*}[t]
\centering
{\small Clear \hspace{18mm} 1,3 \hspace{21mm} 5 \hspace{22mm} 7 \hspace{22mm} 9 \hspace{18mm} I, IA, IB \hspace{14mm} II, III}
\\
\includegraphics[width=.13\textwidth]{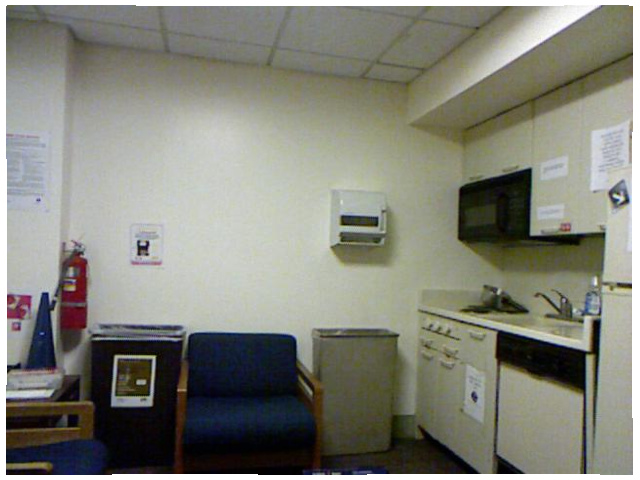}\hfill
\includegraphics[width=.13\textwidth]{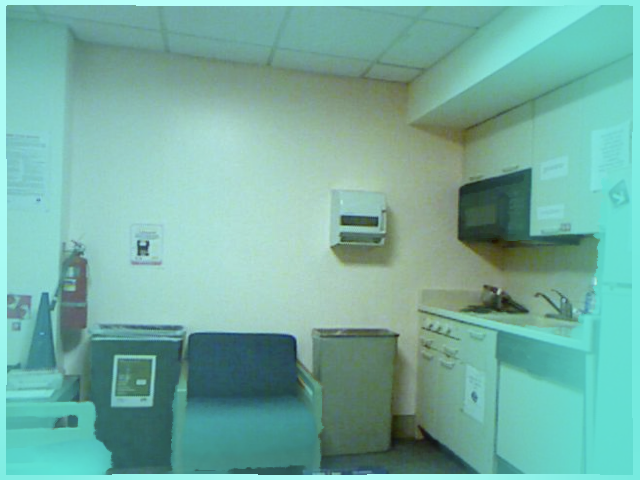}\hfill
\includegraphics[width=.13\textwidth]{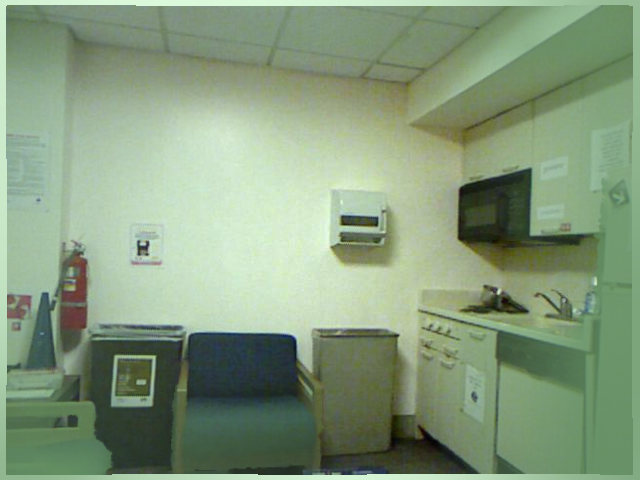}\hfill
\includegraphics[width=.13\textwidth]{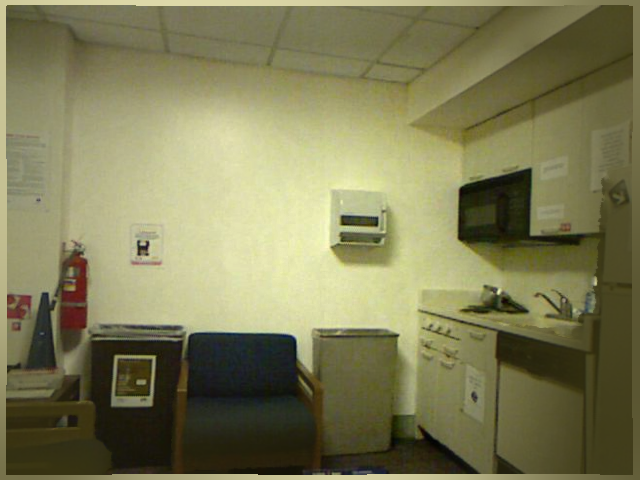}\hfill
\includegraphics[width=.13\textwidth]{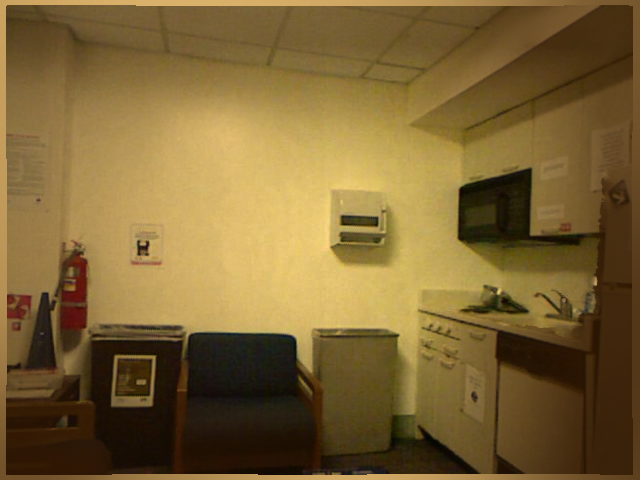}\hfill
\includegraphics[width=.13\textwidth]{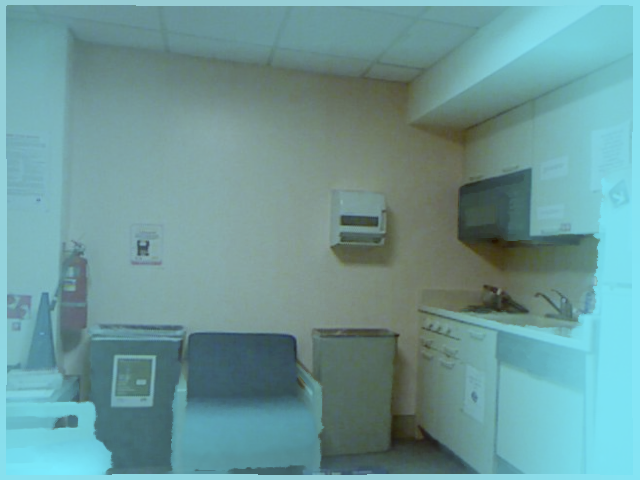}\hfill
\includegraphics[width=.13\textwidth]{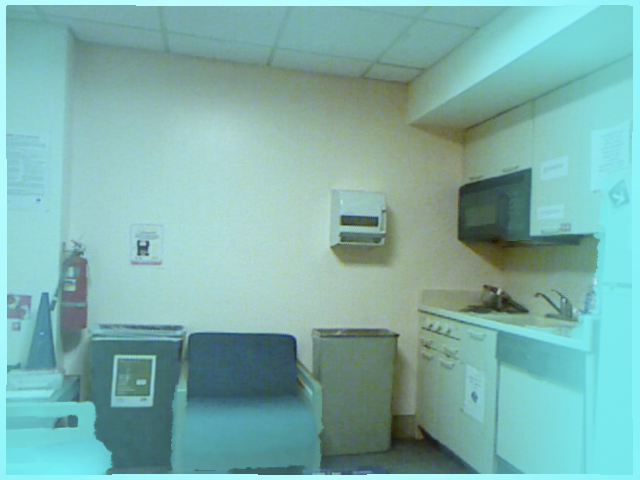}
\\
\caption{Underwater images  synthesized following the approach in \cite{anwar}. We club similar looking water types into a single class and reduce the total number of classes from 10 to 6 in order to boost the performance of our nuisance classifier.}\label{synds6}
\end{figure*}

We follow the approach mentioned in \cite{anwar} to generate synthetic underwater images of multiple water types. The synthetic images are generated using the image formation model described by equations \ref{imgform} and \ref{transcoeff}. We use the NYU-V2 RGB-D dataset \cite{nyudepth} to provide us with the clear images as it also contains the depth information required to generate the corresponding synthetic images. We generate images for 6 Jerlov water types for each image in the dataset instead of generating images for 10 Jerlov water types. We combine similar image types 1 and 3, I, IA and IB and II and III from the 10 Jerlov water types to reduce the proximity between different water types. This boosts the nuisance classifier's performance as it is able to distinguish between different water types more easily. The images are synthesized using different values of $N_{c}$ taken from \cite{anwar} and random $B_{c}$ and $d(x)$ values. For each image in the dataset, and for each of its 6 water types, we augment the dataset by generating 6 images with random $B_{c}$ and $d(x)$ parameters. Thus, for each image in the dataset we have 36 corresponding underwater images of multiple water types. The synthesized 6 types of images for a given image can be seen in figure \ref{synds6}.
\subsubsection{Real-world underwater image dataset}

We use Underwater Image Enhancement Benchmark Dataset (UIEBD) built by \cite{uiebd} as our real-world underwater image dataset. The dataset consists of 890 underwater images.

\subsection{Results on the synthetic dataset}

\subsubsection{Qualitative results}

\begin{figure}[h]
    \centering
        \includegraphics[width=1\linewidth]{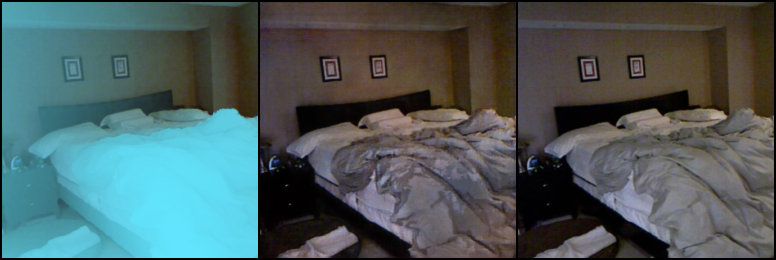}
        \includegraphics[width=1\linewidth]{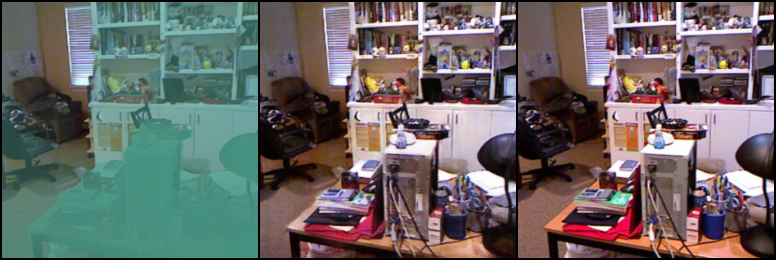}
        \includegraphics[width=1\linewidth]{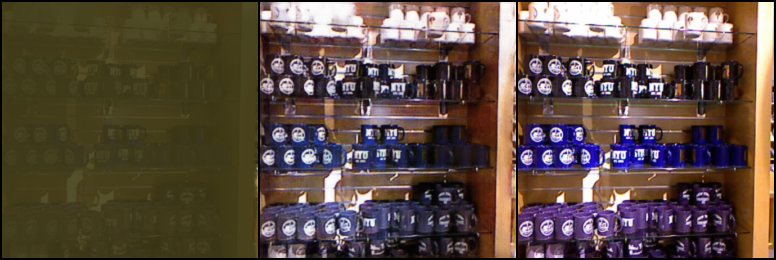}
    \caption{Results on the synthesized underwater dataset. Left column shows the input underwater images, middle column shows the results of our model and the right column shows the ground truth clear images.}\label{uwres}
\end{figure}

Figure \ref{uwres} shows some visual results of our model on the test set of the synthetic underwater dataset which we synthesized in section \ref{synds}. We can visually see that our model is successful in reconstructing the original color of the input images. The output images recover even the minute details from the degraded input images.


\subsubsection{Quantitative results}

We also compute quantitative evaluation metrics like SSIM \cite{ssimcite} and PSNR for the generated images of different Jerlov water types \cite{jerlov} with respect to their clear counterparts. As seen in table \ref{quant_res}, our model outperforms other methods for almost all water types.

\begin{table*}[t]
	\centering
	\begin{tabular}{|l|l|l|l|l|l|l|l|l|l|l|}
		\hline
        \multirow{9}{*}{\textbf{SSIM}} & \textbf{Water Type} & \textbf{RAW} & \textbf{RED} & \textbf{UDCP} & \textbf{ODM} & \textbf{UIBLA} & \textbf{UWCNN} & \textbf{UIE-DAL} \\ \cline{2-9}
         & 1 & 0.7065 & 0.7406 & 0.7629 & 0.724 & 0.6957 & 0.8558 & \multirow{2}{*}{\textbf{0.9313}} \\ \cline{2-8}
         & 3 & 0.5788 & 0.6639 & 0.6614 & 0.6765 & 0.5765 & 0.7951 & \\ \cline{2-9}
         & 5 & 0.4219 & 0.5934 & 0.4269 & 0.6441 & 0.4748 & 0.7266 & \textbf{0.9364} \\ \cline{2-9}
         & 7 & 0.2797 & 0.5089 & 0.2628 & 0.5632 & 0.3052 & 0.607 & \textbf{0.9353} \\ \cline{2-9}
         & 9 & 0.1794 & 0.3192 & 0.1624 & 0.4178 & 0.2202 & 0.492 & \textbf{0.925} \\ \cline{2-9}
         & I & 0.8621 & 0.8816 & 0.8264 & 0.8172 & 0.7449 & \textbf{0.9376} & 0.9129 \\ \cline{2-9}
         & II & 0.8716 & 0.8837 & 0.8387 & 0.8251 & 0.8017 & 0.9236 & \multirow{2}{*}{\textbf{0.9235}} \\ \cline{2-8}
         & III & 0.7526 & 0.7911 & 0.7587 & 0.7546 & 0.7655 & 0.8795 & \\ \hline

        \multirow{8}{*}{\textbf{PSNR}} & 1 & 15.535 & 15.596 & 15.757 & 16.085 & 15.079 & 21.79 & \multirow{2}{*}{\textbf{28.4488}} \\ \cline{2-8}
         & 3 & 14.688 & 12.789 & 14.474 & 14.282 & 13.442 & 20.251 & \\ \cline{2-9}
         & 5 & 12.142 & 11.123 & 10.862 & 14.123 & 12.611 & 17.517 & \textbf{28.6697} \\ \cline{2-9}
         & 7 & 10.171 & 9.991 & 9.467 & 12.266 & 10.753 & 14.219 & \textbf{28.5793} \\ \cline{2-9}
         & 9 & 9.502 & 11.62 & 9.317 & 9.302 & 10.09 & 13.232 & \textbf{27.6551} \\ \cline{2-9}
         & I & 17.356 & 19.545 & 18.816 & 18.095 & 17.488 & 25.927 & \textbf{27.1015} \\ \cline{2-9}
         & II & 20.595 & 20.791 & 17.204 & 17.61 & 18.064 & 24.817 & \multirow{2}{*}{\textbf{28.1602}} \\ \cline{2-8}
         & III & 16.556 & 16.69 & 14.924 & 16.71 & 17.1 & 22.633 & \\ \hline
	\end{tabular}
	\caption{Comparison of our model (UIE-DAL) with SSIM, PSNR values of previous methods. Higher values mean better results. Bold values show the best performer. Values of the previous methods are reprinted from \cite{anwar}.}
	\label{quant_res}
\end{table*}

\subsection{Results on the real-world dataset}

\begin{figure}[h!]
    \centering
        \includegraphics[width=1\linewidth]{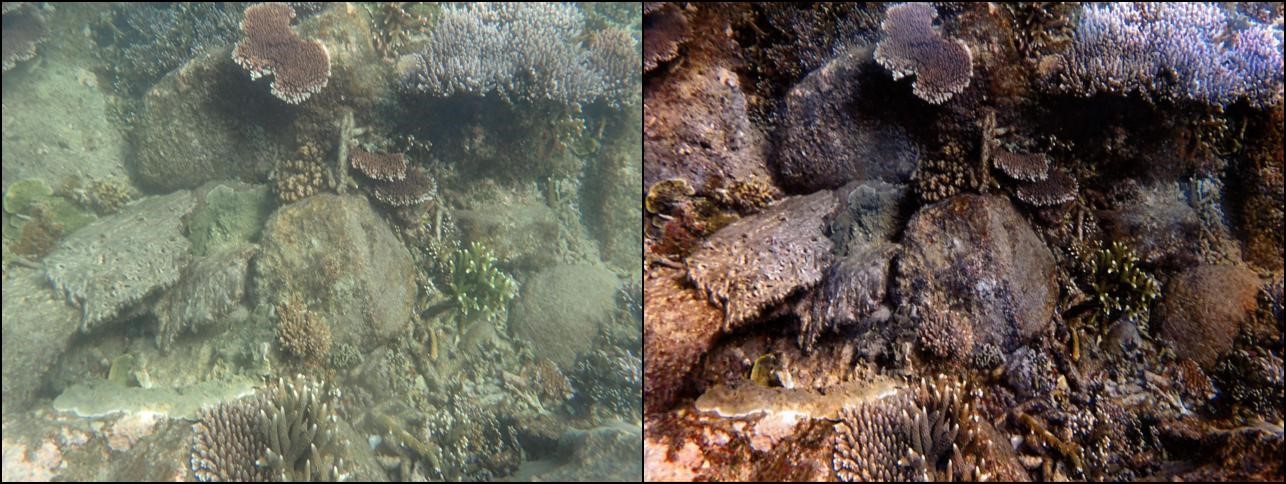}
        \includegraphics[width=1\linewidth]{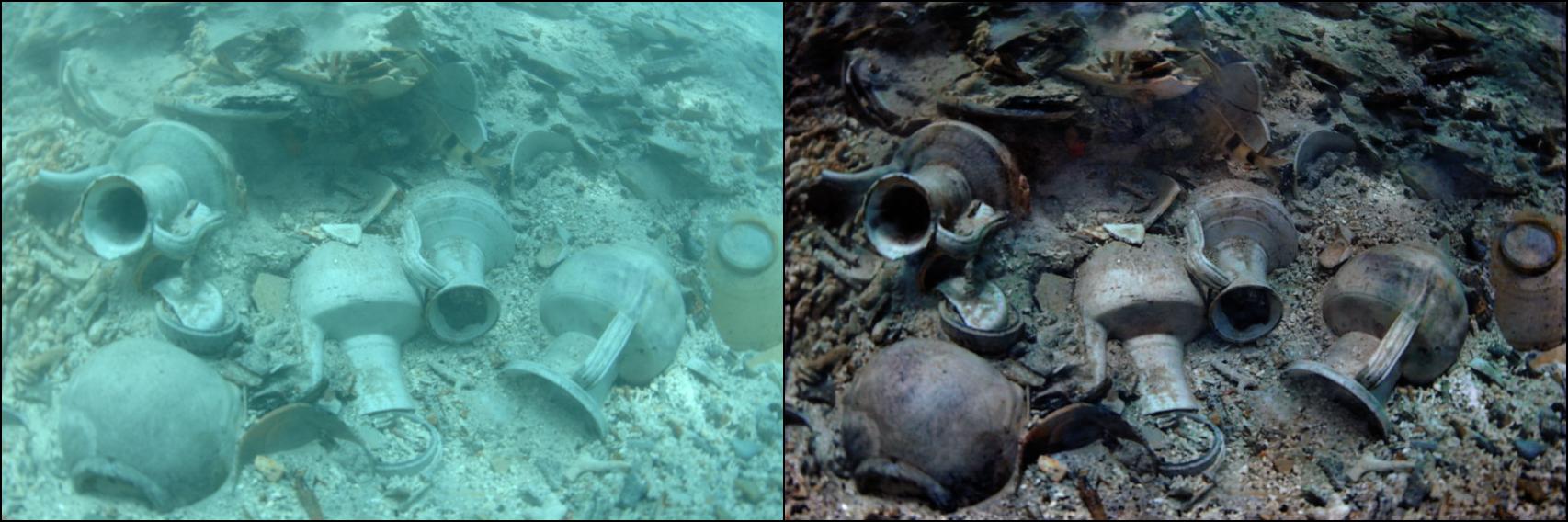}
        \includegraphics[width=1\linewidth]{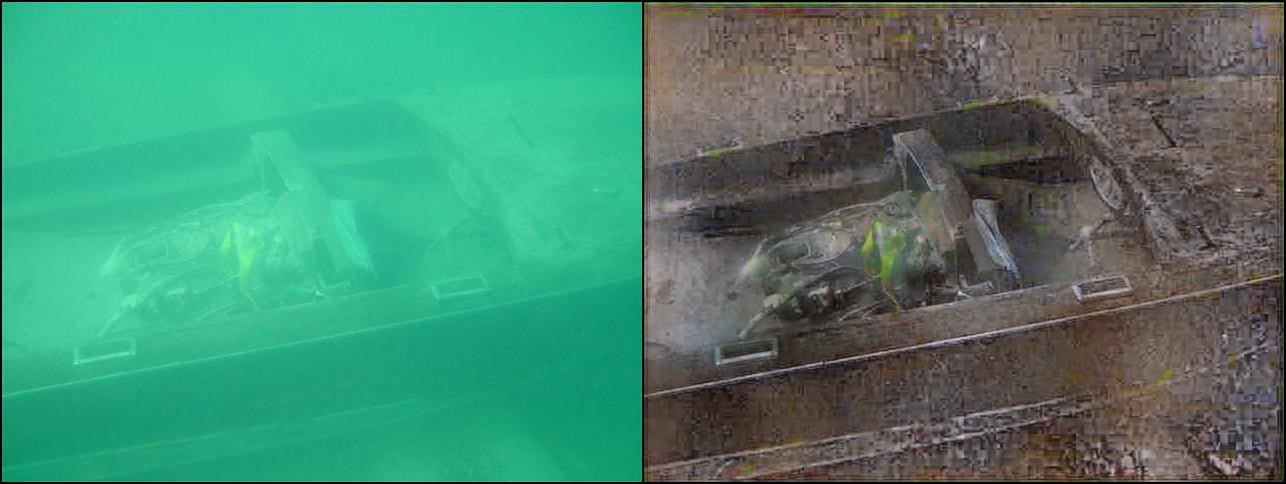}
    \caption{Results on the real-word dataset \cite{uiebd}. Left column shows the input underwater images and the right column shows the results of our model.}\label{rwres}
\end{figure}

We also test our model on a real-world dataset to see the transferability of our model to different datasets. Figure \ref{rwres} shows some visual results of our model on the Underwater Image Enhancement Benchmark Dataset \cite{uiebd}. Here, we see that the model performs well and is able to generalize on image distributions different than that of the training images. Handling such diversity is one of our main goals apart from generating clear underwater images.  

\subsection{Comparison to no adversarial loss}

\begin{figure}[h!]
    \centering
        \includegraphics[width=\linewidth]{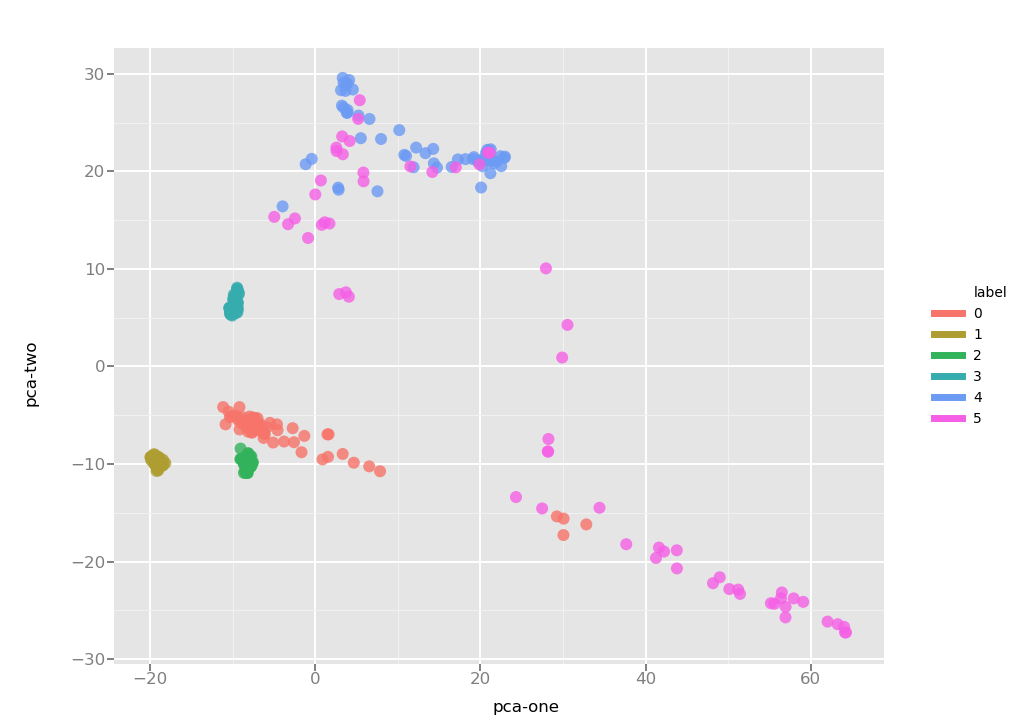}
            {\small (a)}
        \includegraphics[width=\linewidth]{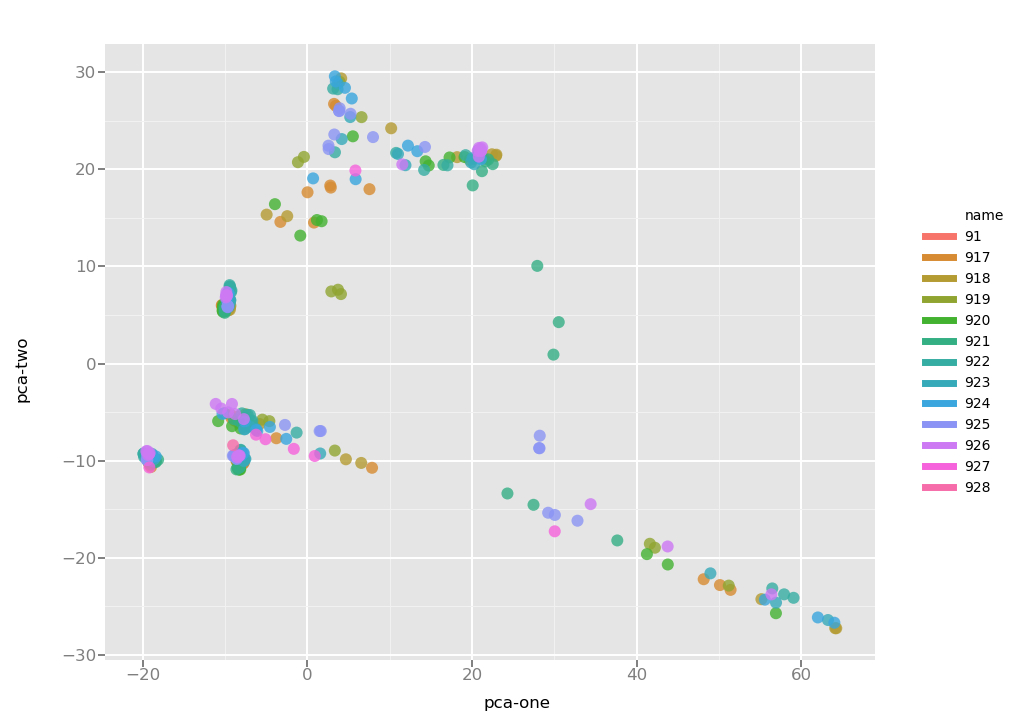}
            {\small (b)}
    \caption{Visualizing first two PCA components of the encoding $Z$ learned by U-Net without adversarial loss. (a) Colors points with the same water type, (b) Colors points with the same content.}\label{pcaunet}
\end{figure}

\begin{figure}[h!]
    \centering
        \includegraphics[width=\linewidth]{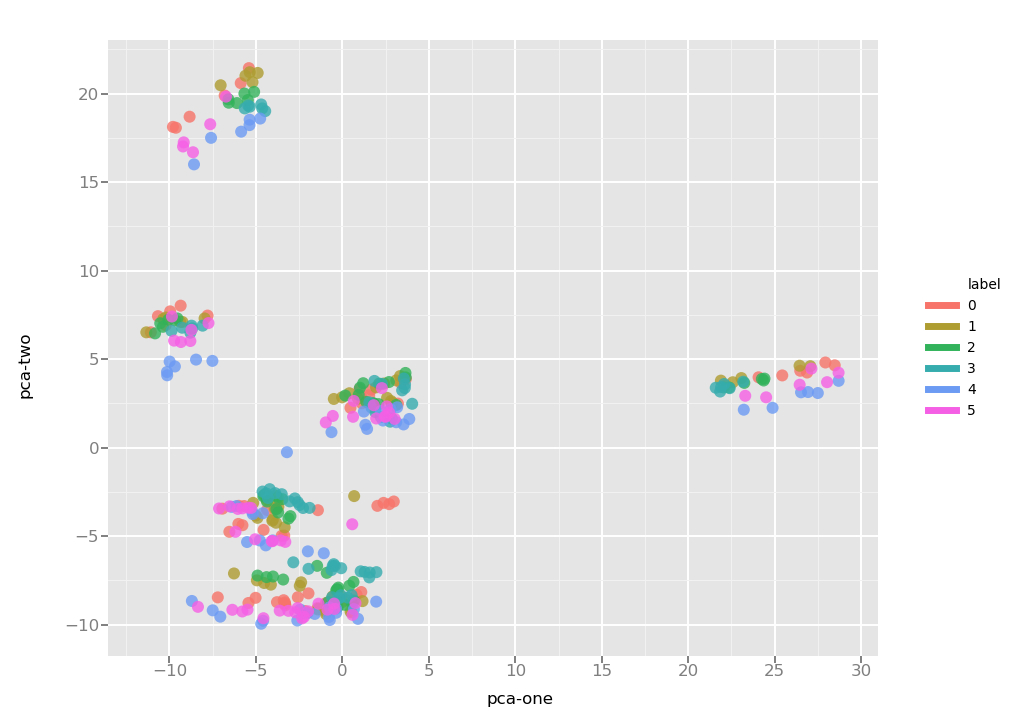}
            {\small (a)}
        \includegraphics[width=\linewidth]{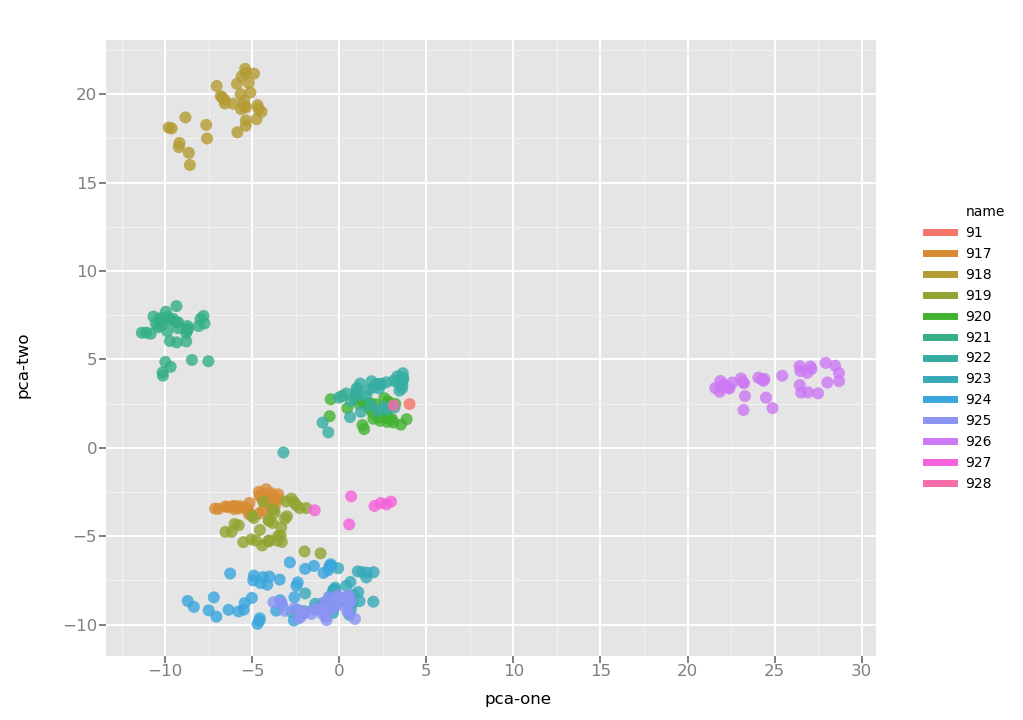}
            {\small (b)}
    \caption{Visualizing first two PCA components of the encoding $Z$ learned by U-Net with adversarial loss (UIE-DAL). (a) Colors points with same water type, (b) Colors points with same content.}\label{pcaunetadv}
\end{figure}

We compare our model with vanilla U-Net without the adversarial loss. To see if we have learned the domain agnostic features, we plot the first two principal components of the encoding $Z$ from both the vanilla U-Net and U-Net with the adversarial loss. We color the points once by the water types and once by the image content for the same set of images. The plotted PCA components can be seen in figures \ref{pcaunet} and \ref{pcaunetadv} respectively.

\begin{figure}[t]
    \centering
        \includegraphics[width=0.95\linewidth]{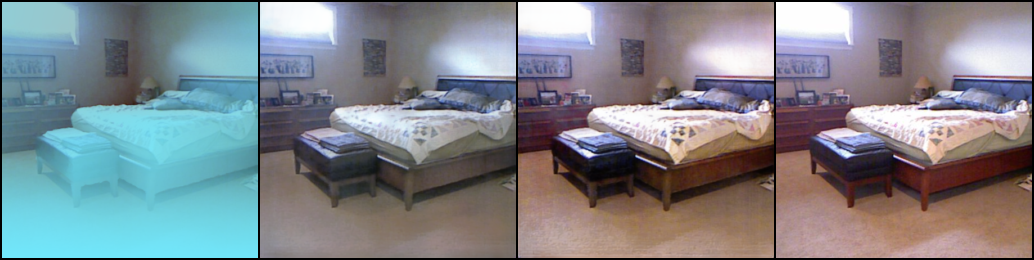}
        {\small (a) From left to right - Input image, output of vanilla U-Net, output of U-Net with adversarial loss, ground truth image.}
        \label{c_sd}
        \includegraphics[width=0.95\linewidth]{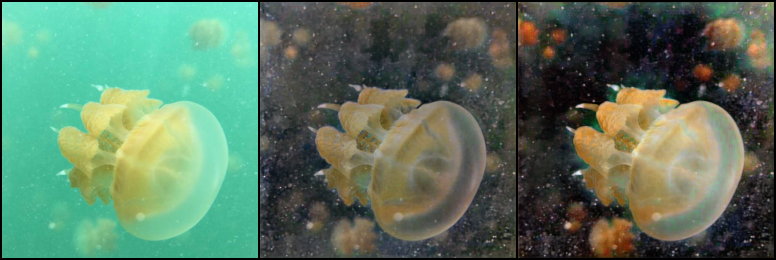}
        {\small (b) From left to right - Input image, output of vanilla U-Net, output of U-Net with adversarial loss.}
        \label{c_rd}
	\caption{Comparison of U-Net with and without adversarial loss. (a) shows results on synthetic data, where as (b) shows results on real-world data.}\label{unvunadv}
\end{figure}

\begin{figure}[h!]
    \centering
        \includegraphics[width=0.45\linewidth]{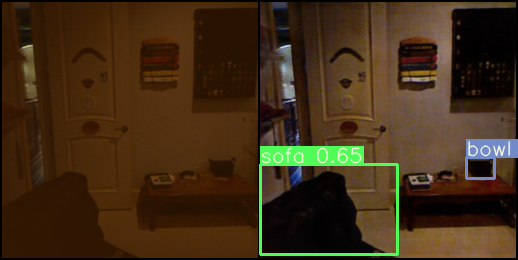}
        \includegraphics[width=0.45\linewidth]{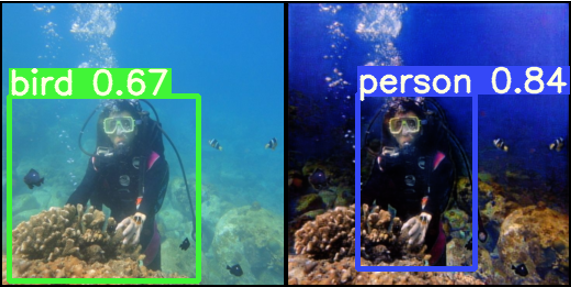}
        \\
        \includegraphics[width=0.45\linewidth]{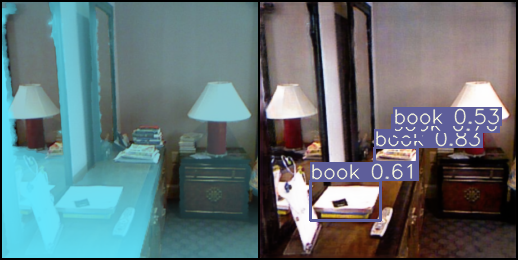}
        \includegraphics[width=0.45\linewidth]{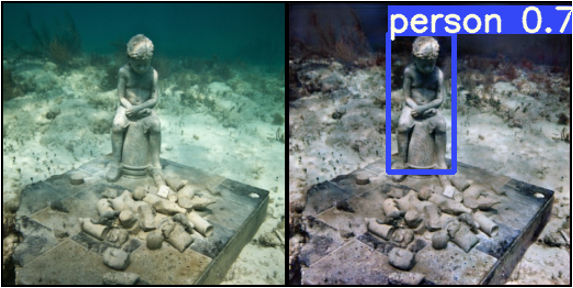}
        \\
         \hspace{1cm}{\small (a) \hfill (b)\hfill (c)\hfill (d)}\hspace{1cm}
        \label{od_rd}
	\caption{Object detection results before and after enhancement (a) Synthetic underwater image, (b) Output of our model for the synthetic underwater image, (c) Real-world underwater image and (d) Output of our model for the real-world image.}\label{objdet}
\end{figure}

\begin{table}
	\centering
	\begin{tabular}{|l|l|l|l|l|l|l|l|l|l|l|}
		\hline
        \multirow{9}{*}{\textbf{SSIM}} &  \textbf{Water Type} & \textbf{U-Net} & \textbf{UIE-DAL (Ours)} \\ \cline{2-4}
         & 1 & \multirow{2}{*}{0.8691} & \multirow{2}{*}{\textbf{0.9313}} \\ \cline{2-2}
         & 3 & & \\ \cline{2-4}
         & 5 & 0.8733 & \textbf{0.9364} \\ \cline{2-4}
         & 7 & 0.8687 & \textbf{0.9353} \\ \cline{2-4}
         & 9 & 0.8614 & \textbf{0.925} \\ \cline{2-4}
         & I & 0.8385 & \textbf{0.9129} \\ \cline{2-4}
         & II & \multirow{2}{*}{0.8385} & \multirow{2}{*}{\textbf{0.9235}} \\ \cline{2-2}
         & III & & \\ \hline

        \multirow{9}{*}{\textbf{PSNR}} & 1 & \multirow{2}{*}{21.6283} & \multirow{2}{*}{\textbf{28.4488}} \\ \cline{2-2}
         & 3 & & \\ \cline{2-4}
         & 5 & 22.6119 & \textbf{28.6697} \\ \cline{2-4}
         & 7 & 22.5754 & \textbf{28.5793} \\ \cline{2-4}
         & 9 & 22.5263 & \textbf{27.6551} \\ \cline{2-4}
         & I & 22.3236 & \textbf{27.1015} \\ \cline{2-4}
         & II & \multirow{2}{*}{21.8279} & \multirow{2}{*}{\textbf{28.1602}} \\ \cline{2-2}
         & III & & \\ \hline
	\end{tabular}
	\caption{Our comparison with SSIM, PSNR values of U-Net without adversarial loss. Higher values mean better results. Bold values show the best performer.}
	\label{t_uvua}
\end{table}

It can be seen from figures \ref{pcaunet} and \ref{pcaunetadv} that we are indeed learning domain agnostic features using adversarial loss. The encoding $Z$ is clustered by the water types in vanilla U-Net, whereas it is clustered by the image content in U-Net with adversarial loss.

We also visually and quantitatively compare both the models. Figure \ref{unvunadv} shows us the visual results of the models on both the synthetic underwater image dataset and the real-world UIEBD. Table \ref{t_uvua} shows us the quantitative comparison.

We can see from both figure \ref{unvunadv} and table \ref{t_uvua} that U-Net with adversarial loss outperforms vanilla U-Net. U-Net with adversarial loss is able to learn domain agnostic features and hence also generates images with rich color quality than vanilla U-Net.

\subsection{Object detection on enhanced images}

As advocated by many previous works \cite{wang2016studying,li2017aod,liu2017image,liu2017enhance,li2018end,liu2018improved,li2019benchmarking,vidalmata2019bridging,li2019single}, the high-level computer vision performance (such as object detection) on enhanced images could act as an indicator of the image enhancement performance itself. We run object detection experiments on the images generated by our model to see if they can help in different underwater vision tasks. We run YOLO v3 \cite{yolov3} object detector on the degraded underwater images and their enhanced versions generated by our model. We observe that object detection is better on the images generated by our model compared to the degraded underwater images of the synthesized underwater dataset. However, we get mixed results when we run the object detector on the real-world UIEBD. Figure \ref{objdet} shows the results of YOLO v3 before and after processing the images with our model.

\section{Conclusion}

We are able to provide a novel solution for underwater image enhancement which outperforms the previous methods both qualitatively and quantitatively. Our goal is to provide a generalized solution which could handle the diversity of the underwater images as well as transform them into clear images. Our model is successful in doing so by learning domain agnostic features for multiple underwater image types and then generating their clear version from those features. We also show that the model is able to generalize well on the unseen real-world data. Also, experimental results on object detection task show that enhancing underwater images with our model before high level vision tasks improves the detection performance. 

{\small
\bibliographystyle{unsrt}
\bibliography{egbib}
}

\end{document}